\documentclass[letterpaper, 10pt, conference]{ieeeconf}
\IEEEoverridecommandlockouts    % to override locked commands
\usepackage{multirow}
\usepackage{lipsum}
\usepackage{amsmath}
\usepackage{amssymb}
\usepackage[ruled,vlined]{algorithm2e}
\usepackage{graphicx}
\graphicspath{{./Figures/}}
\usepackage[caption=false,font=footnotesize]{subfig}
\usepackage{url}
\usepackage{cite}
\usepackage[hidelinks]{hyperref}
\title{\LARGE \bf Learning from Shared-Control Overrides: Context-Driven Acceleration Profile Prediction for Personalized Overtaking}

\author{
	\parbox{\textwidth}{%
		\centering
		Ruizheng Xu$^{1, 2}$, Lounis Adouane$^{1}$, Javier Ibañez-Guzmán$^{2}$, Clément Zinoune$^{2}$%
	}%
	\thanks{$^{1}$ Heudiasyc UMR CNRS 7253, Université de Technologie de Compiègne (UTC), Compiègne, France.
		{\tt\small \{ruizheng.xu, lounis.adouane\}@hds.utc.fr}}%
	\thanks{$^{2}$Ampere Software Technology, Renault Group, Guyancourt, France.
		{\tt\small \{ruizheng.xu, javier.ibanez-guzman, clement.c.zinoune\}@ampere.cars}}%
}

\begin{document}
	
	\maketitle
	\thispagestyle{empty}
	\pagestyle{empty}
	
	%%%%%%%%%%%%%%%%%%%%%%%%%%%%%%%%%%%%%%%%%%%%%%%%%%%%%%%%%%%%%%%%%%
	\begin{abstract}
        Adaptive Cruise Control (ACC) systems are typically calibrated for an average driver, often resulting in a mismatch between vehicle behavior and individual expectations during time-critical maneuvers such as highway overtaking. When the ACC is perceived as too conservative and inconsistent, drivers intervene through throttle overrides, providing implicit feedback on the system's behavior.

        This paper reframes these override actions as human-in-the-loop supervisory signals and proposes a data-driven framework for personalized vehicle adaptation, termed Context-driven Personalized ACC (CoP-ACC). Rather than relying solely on end-to-end regression, which tends to over-smooth dynamic responses, we introduce a hybrid pipeline combining: (i) unsupervised hierarchical clustering to extract representative acceleration profiles from override events; (ii) a context classifier that maps pre-maneuver driving conditions to the appropriate profile; and (iii) a residual regressor that refines the selected profile into a smooth, personalized acceleration profile tailored to the immediate context.
        
        Evaluated on real-world public-road data against a withheld forced-ACC baseline, the approach demonstrates high reconstruction fidelity and generates acceleration profiles that tend toward the driver's expected behavior in potential override contexts. The results highlight the potential of learning from shared-control overrides to enable anticipatory, personalized ACC behavior, reducing manual interventions and improving ride comfort.
	\end{abstract}
	
	%%%%%%%%%%%%%%%%%%%%%%%%%%%%%%%%%%%%%%%%%%%%%%%%%%%%%%%%%%%%%%%%%%
	\section{Introduction}
	\label{sec:introduction}

    Adaptive Cruise Control (ACC) significantly reduces driver workload \cite{aliAdaptiveCruiseControl2024}, but its ``one-size-fits-all" calibration limits user acceptance during complex highway maneuvers. Notably, drivers commonly keep ACC engaged during overtaking. We define this scenario as an \textit{ACC-assisted overtake}: the driver manually initiates and steers the lane change, while the ACC remains active and automatically controls longitudinal acceleration to manage speed relative to the overtaken vehicle. Because drivers possess highly individualized driving tolerances, generic ACC systems often execute conservative accelerations that feel delayed or insufficient \cite{wangPersonalizedAdaptiveCruise2021}. This human-machine discrepancy forces the driver to intervene and physically override the pedal to achieve the desired overtake urgency \cite{maDriversTrustAcceptance2021}.

    Recent ACC personalization studies \cite{hasenjagerSurveyPersonalizationAdvanced2020} predominantly target steady-state car-following or rely on end-to-end continuous regression. Applied to highly variable driving data where even a single individual driver exhibits significantly different behaviors depending on their immediate situational context \cite{liaoReviewPersonalizationDriving2024}, these regressors suffer from ``regression to the mean," producing over-smoothed profiles that fail to capture urgency during transient maneuvers like overtaking. Furthermore, end-to-end models lack interpretability for functional safety. Finally, although some studies have begun using driver overrides to adapt gap preferences in car-following \cite{zhao_real-time_2023}, none exploit them to learn continuous, context-dependent acceleration profiles for transient maneuvers like overtaking.  
    
    To bridge this gap, this paper proposes CoP-ACC, a Context-Driven Personalized ACC framework treating manual interventions as explicit ground-truth labels of driver expectation, formally defined as shared control override events. Targeting the overtaking scenario, we introduce a hybrid machine-learning pipeline to overcome regression limitations. Unsupervised hierarchical clustering extracts discrete acceleration profiles, preserving dynamics. This intent is mapped to pre-maneuver kinematics using supervised classification (Random Forest), ensuring interpretability. Finally, a Context-Conditioned 1D-CNN Decoder generates continuous residual tracking modifications. We introduce a distributional evaluation methodology, proving the framework's ability to preemptively fulfill individual driver expectations on withheld data.

    This paper is organized as follows: Section II reviews related work in ADAS personalization. Section III outlines the proposed methodology. Section IV details the data acquisition and experimental protocol. Section V presents the experiments and results. Section VI concludes the paper and discusses future works.
	
    %%%%%%%%%%%%%%%%%%%%%%%%%%%%%%%%%%%%%%%%%%%%%%%%%%%%%%%%%%%%%%%%%%
	\section{Related Works}
	\label{sec:relatedworks}

    This section reviews two bodies of work relevant to the proposed framework: personalization techniques for longitudinal vehicle control in Section \ref{subsec:personalization_longitudinal} and the current state of overtaking automation and shared control in Section \ref{subsec:overtaking_sharedcontrol}.
	
	\subsection{Personalization in Longitudinal Control}
    \label{subsec:personalization_longitudinal}

    User acceptance of Advanced Driver Assistance Systems (ADAS) is heavily dependent on how closely the machine's behavior aligns with an individual driver's internal expectations \cite{hasenjagerSurveyPersonalizationAdvanced2020}. To address the limitations of rigid traditional control (e.g., MPC, PID), recent literature has shifted toward data-driven longitudinal personalization, aiming to deduce and replicate these individualized preferences. 

    One primary approach to individual personalization relies on style categorization. Techniques such as unsupervised clustering and Inverse Reinforcement Learning (IRL) are used to extract individualized driving styles from traffic data \cite{gaoPersonalizedAdaptiveCruise2020, shengStudyLearningSimulating2022, zhaoPersonalizedCarFollowing2022}. For instance, \cite{gaoPersonalizedAdaptiveCruise2020} successfully employs clustering and classification to identify driver styles online, but relies on traditional MPC for longitudinal control. Similarly, \cite{zhao_real-time_2023} use IRL to dynamically update scalar gap preferences. While these methods successfully categorize individual behaviors, two main limitations are: an exclusive focus on steady-state gap maintenance (car-following) and reliance on heuristic physical controllers for output. They do not address the highly transient, aggressive dynamics required for maneuver-specific accelerations, such as overtaking.
    
    Another approach achieves more nuanced individual control by formulating personalization as a continuous trajectory prediction problem. Both works from \cite{wangGaussianProcessBasedPersonalized2022} and \cite{lefevreLearningBasedFrameworkVelocity2016} utilize methods such as Gaussian Process Regression (GPR) and Gaussian Mixture Regression (GMR) to predict acceleration directly from environmental states. While effective, treating individual personalization exclusively as an end-to-end regression task presents two critical limitations. First, by skipping an explicit style categorization pipeline, these models naturally converge toward overly smoothed average trajectories (``regression to the mean"). Because human driving data embodies both gentle and aggressive reactions, a single continuous regressor allows to smooth out these extremes. Consequently, these models often fail to execute the decisive, high-urgency acceleration peaks unique to individual drivers—the sudden surges in speed strictly required to safely clear a slower vehicle during overtaking. Second, when predicting trajectories using complex, end-to-end models like Artificial Neural Networks \cite{navaPersonalizedAdaptiveCruise2019}, the system functions as a ``black box". This lack of interpretability makes it mathematically impossible to attribute exactly which environmental feature triggered a specific acceleration output, posing a significant challenge for safety compliance in modern ADAS.
    
    The proposed CoP-ACC methodology overcomes these limitations by introducing a novel, hybrid machine-learning pipeline designed explicitly for individual driver personalization (cf. Section \ref{sec:methodology}). By utilizing unsupervised clustering to extract discrete acceleration profile templates prior to supervised classification and continuous residual regression, the proposed framework prevents over-smoothing and preserves the unique dynamics of each driver based on its situational context. Furthermore, the classification stage employs an interpretable Random Forest whose feature contributions are quantified through SHapley Additive exPlanations (SHAP) \cite{lundbergUnifiedApproachInterpreting2017} analysis, ensuring transparent mapping between pre-maneuver environmental features and the predicted driver intent. This approach effectively bridges the gap between discrete driving styles and continuous profile prediction, providing a solution specifically tailored for the overtaking maneuver.
    
    \subsection{Overtaking and Shared Control}
    \label{subsec:overtaking_sharedcontrol}
    
    Overtaking is a complex, multi-phase maneuver \cite{dozzaHowDriversOvertake2016} that presents a significant challenge for generalized ADAS. When a generic ACC performs a conservative acceleration that fails to match the driver's expectations, the human-machine mismatch prompts a shared control override via the pedals \cite{marcanoCanSharedControl2022}. While works like \cite{marcanoReviewSharedControl2020} extensively study shared control during overtaking, their focus remains predominantly on lateral steering conflicts and safety assessments rather than learning the personalized longitudinal acceleration style. Additionally, while recent literature has begun treating overrides as implicit feedback, they are generally used to incrementally tune scalar parameters such as gap preferences \cite{zhao_real-time_2023}, rather than to learn the continuous acceleration profile expected by the driver during a specific transient maneuver.

    Currently, there is a distinct lack of frameworks that extract the full, continuous kinematic trajectory of a shared control override to use as a primary teaching signal for overtaking maneuvers. This paper addresses this gap by exclusively utilizing these real-world shared-control conflicts to drive the aforementioned hybrid personalization pipeline. By treating shared control overrides as ground-truth acceleration profiles rather than simple post-hoc parameter tuners, the proposed framework can preemptively output the precise acceleration curve required to resolve the human-machine conflict during an overtake before it arises.
    
    %%%%%%%%%%%%%%%%%%%%%%%%%%%%%%%%%%%%%%%%%%%%%%%%%%%%%%%%%%%%%%%%%
	\section{CoP-ACC: Context-Driven Personalized ACC Framework}
	\label{sec:methodology}

    This section first describes the ACC operating principle and the shared control mechanism exploited by CoP-ACC, then details the framework architecture, the unsupervised profiling stage, pre-maneuver feature extraction, and the context-driven classification and residual regression pipeline.

    \subsection{ACC Operating Principle}

    ACC operates in two modes: cruise mode, maintaining a driver-set target speed, and following mode, adjusting speed to preserve a safe gap behind a target vehicle. Driver pedal inputs govern state transitions: pressing the throttle pedal triggers a temporary override allowing manual acceleration, and on pedal release the ACC automatically resumes its target; pressing the brake pedal immediately deactivates the system and returns full control to the driver \cite{yuResearchesAdaptiveCruise2022}. We define this driver intervention as a shared control override. During overtaking, the ACC remains engaged in following mode. However, its conservative acceleration often falls short of the driver's intended acceleration, resulting in a throttle override that constitutes the shared control signal studied in this work.

    \subsection{Framework Overview}

    \begin{figure*}[t]
        \centering
        \includegraphics[width=.92\textwidth]{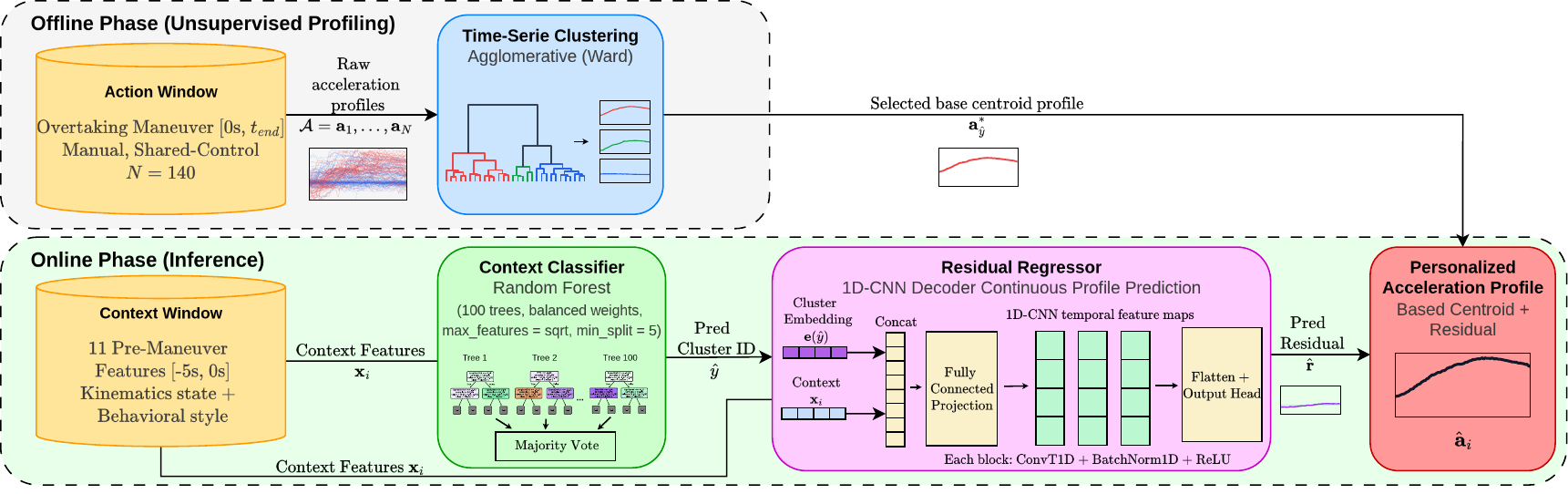}
        \caption{Overview of the proposed CoP-ACC framework: offline unsupervised profiling and online context-driven prediction pipeline.}
        \label{fig:framework}
    \end{figure*}

    We propose a hybrid, context-driven architecture based on machine learning methods (cf Fig. \ref{fig:framework}). The framework operates in two primary stages: offline unsupervised profiling of driver behavioral expectations, and online supervised classification and regression to predict the appropriate personalized acceleration profile.

    As shown in Fig. \ref{fig:framework}, this architecture is separated into an offline training phase and an online inference phase. During offline training, the pipeline proceeds in three steps: (1) raw acceleration traces from overtaking events are clustered via time-series clustering to identify distinct driver behavioral preferences, producing $K$ base centroid profiles; (2) each event is assigned a cluster label, and a context classifier is trained to predict this label from an 11-dimensional pre-maneuver kinematic and behavioral feature vector; (3) a residual regressor is trained to predict the individual residual between each event's actual profile and its assigned centroid, conditioned on both the context features and the predicted cluster label.

    During online inference, the system receives a new pre-maneuver context vector and executes the trained pipeline: the classifier predicts the appropriate cluster, the regressor generates the corresponding residual curve, and the final personalized acceleration profile is computed as the sum of the selected base centroid and the predicted residual. The concrete data collection protocol, temporal windowing, and event annotation process are detailed in Section IV.

    \subsection{Unsupervised Acceleration Profile Clustering}

    The first phase extracts a discrete set of behavioral profiles from the uncorrupted dataset, which strictly comprises naturalistic manual driving and shared control override events. We explicitly exclude passively accepted ACC events, as only active pedal interventions contain the true ground-truth of the driver's desired acceleration. This is framed as a time-series clustering problem.

    We employ Agglomerative Hierarchical Clustering \cite{lanceGeneralTheoryClassificatory1967} utilizing Euclidean distance on the set of normalized acceleration profiles $\mathcal{A} = \{\mathbf{a}_1, \dots, \mathbf{a}_N\}$ with Ward's minimum variance criterion to ensure dense, spherical clusters \cite{wardjr.HierarchicalGroupingOptimize1963}.
    
    To ensure the framework remains scalable across different individual drivers, the number of behavioral profiles ($K$) is automatically determined via a Silhouette Stability Analysis \cite{rousseeuwSilhouettesGraphicalAid1987}. This method calculates Silhouette Scores across a range of $K$, evaluating the performance drop between successive values. The algorithm dynamically selects the first $K$ where the gap to $K+1$ drops below a predefined stability threshold, indicating diminishing returns in structural distinction. For our subject, optimal separation stabilized at $K=3$. The resulting clusters, $C_k$ where $k \in \{0, 1, 2\}$, are defined by their base centroid acceleration profile $\mathbf{a}^*_k$ (cf. Fig. \ref{fig:cluster_profiles}).

    \subsection{Pre-Maneuver Feature Extraction}

    To predict the expected profile entirely from pre-maneuver conditions \cite{bouhsissinDriverBehaviorClassification2023}, we extracted 11 state variables to populate the context vector $\mathbf{x}_i$. These features comprise two categories: Kinematic State features including Ego Speed, Relative Speed, Distance to Target, Speed Deficit to Target, Time Headway, and Inverse Time-To-Collision evaluated at $t=0$; and Behavioral Style features including standard deviations of steering angle, longitudinal jerk, and throttle, along with maximum lateral jerk and mean target acceleration aggregated over the $[-5s, 0s]$ context pre-maneuver window (cf. Fig. \ref{fig:framework}).

    \subsection{Context-Driven Classification and Residual Regression}

    With the context vectors $\mathbf{x}_i$ and corresponding cluster labels $y_i \in \{0, 1, 2\}$ established, we employ a Random Forest classifier \cite{breimanRandomForests2001} to map the pre-maneuver context to a discrete base profile $\hat{y}$. Both classifier and residual regressor use an 80/20 stratified split, and implementation-level model settings are summarized in Fig. \ref{fig:framework}. Both models are trained on the combined dataset of naturalistic manual driving and shared control override events, allowing them to learn the driver's underlying preference through both how they naturally execute overtaking maneuver and how they explicitly override the system.

    However, mapping to a single rigid centroid only provides an averaged expectation based on the categorical cluster. It fails to capture the fluid continuity of human driving or adjust for inner-cluster variance. Therefore, we introduce a Continuous Profile Prediction strategy via Residual Regression. Recognizing that the driver's ultimate input is a modification of the base intent, we compute the residual curve as:
    \begin{equation}
        \mathbf{r}_i = \mathbf{a}_i - \mathbf{a}^*_{y_i}
    \label{eq:residual_curve}
    \end{equation}

    A context-conditioned 1D-CNN Decoder is trained to predict this continuous residual curve $\mathbf{\hat{r}}$. The decoder conditions on both the 11-dimensional context vector and the predicted cluster label. The cluster label is represented through a learned embedding. The combined conditioning information is then mapped through fully connected layers and transposed one-dimensional convolutions to generate a fixed-length 100-point residual acceleration profile. The final generated acceleration profile is the summation of the base centroid and the predicted residual. This hybrid approach preserves the discrete behavioral intent while injecting the nuanced, continuous adjustments demanded by the real-time context.
    
    %%%%%%%%%%%%%%%%%%%%%%%%%%%%%%%%%%%%%%%%%%%%%%%%%%%%%%%%%%%%%%%%%%
	\section{Data Acquisition and Processing}
	\label{sec:data_acquisition}

    This section describes the experimental setup, data collection protocol, and the curation pipeline used to prepare the training dataset.

    \subsection{Experimental Setup}

\begin{figure}[!htbp]
        \centering
        % First subfigure
        \subfloat[]{%
            \includegraphics[height=2.8cm]{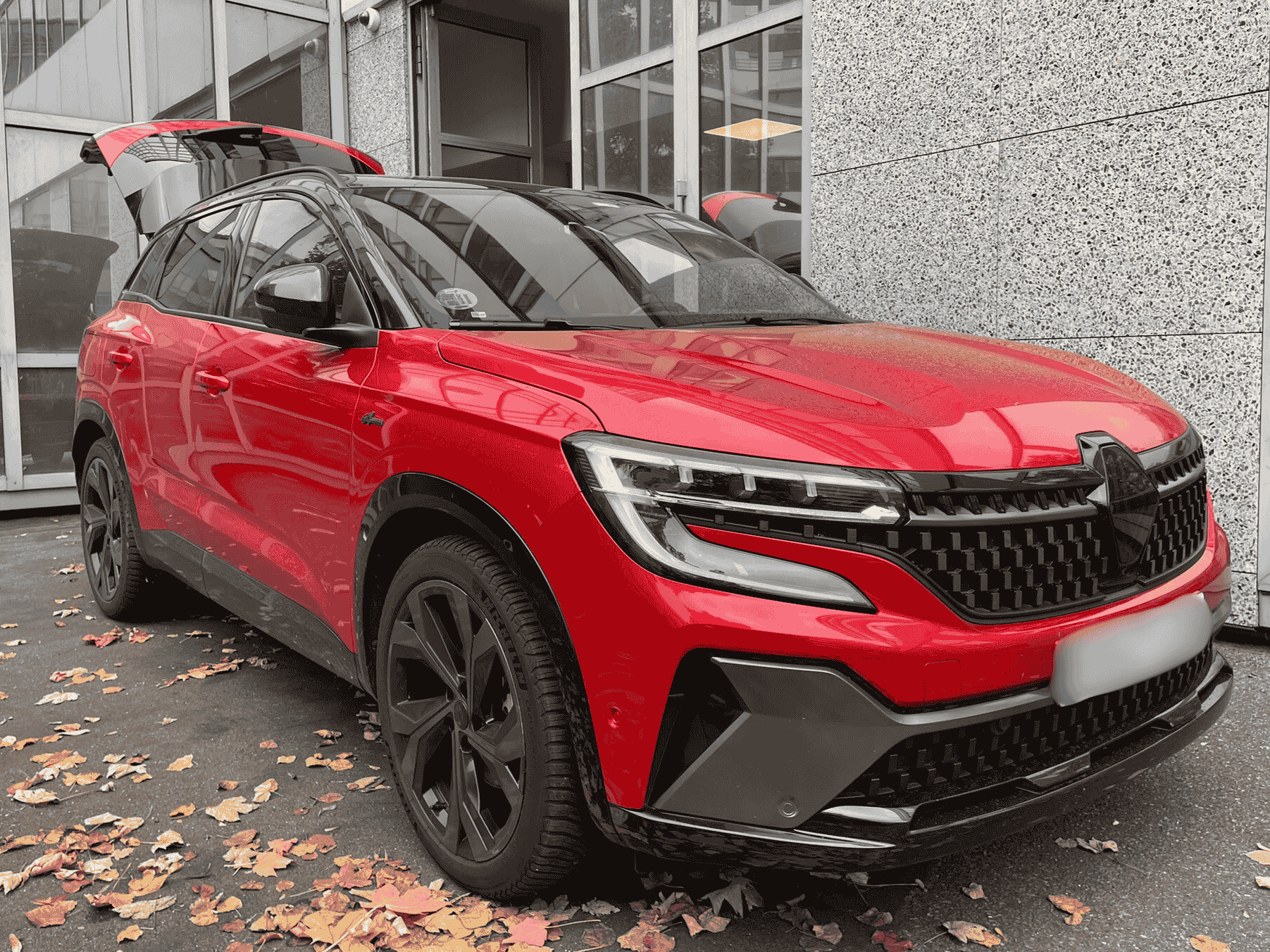}%
        \label{fig:australe}}
        \hfil
        % Second subfigure
        \subfloat[]{%
            \includegraphics[height=2.8cm]{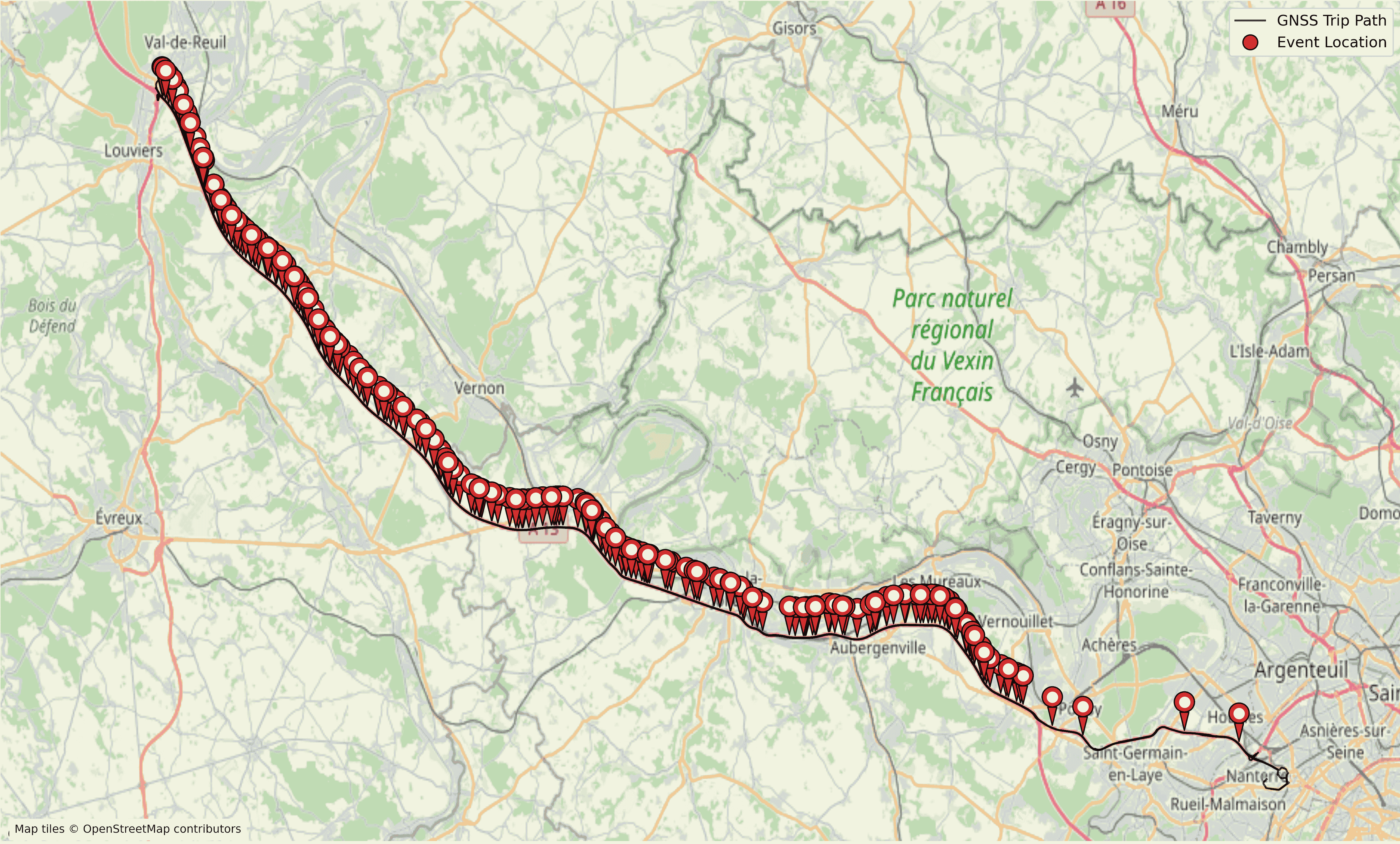}%
        \label{fig:trip}}
        
        \caption{(a) Instrumented vehicle used for data collection. (b) GNSS path of round trips with annotated overtaking events.}
        \label{fig:experiement_setup}
    \end{figure}

    Data were logged directly from an instrumented vehicle's (Renault Austral cf. Fig \ref{fig:australe}) Controller Area Network (CAN) bus with production-level ADAS. Four primary signal categories were identified from the telemetry: Vehicle States characterizing ego and target kinematics (e.g., longitudinal/lateral acceleration, relative speed, distance); Driver Inputs \& HMI detailing physical interventions and statuses (e.g., throttle position, steering angle, ACC overrides); Environmental Context defining the tactical driving corridor (e.g., lane positioning, speed limits).

    To achieve individual-level ADAS personalization, this study deliberately adopts a single-subject data collection approach for this initial framework, aiming to demonstrate personalization for one specific driver before scaling up. The subject is an eight-year experienced highway commuter with a dynamic driving style. While this provides the isolated signals necessary to establish the methodology, we plan future data collection across diverse driver profiles to validate generalizability.

    The data collection procedure was conducted on a three-lane highway in France during low-traffic periods. The driver completed three predefined round trips as shown in Figure \ref{fig:experiement_setup}, each capturing a distinct driving paradigm. In Trip 1 (Forced-ACC Baseline), the vehicle was entirely controlled by the ACC and the driver was strictly forbidden from overriding via the pedals, establishing a reference dataset of the ACC's generic acceleration profile. In Trip 2 (Naturalistic Manual), the ACC was completely disabled and the driver operated the vehicle naturally during overtakes. In Trip 3 (Shared-Control), the ACC was active, but the driver freely pressed the accelerator pedal to override the system whenever its longitudinal behavior failed to meet their personal expectations, providing a ground-truth shared control override acceleration profile.
    
    \subsection{Data Curation and Event Annotation}

    All raw vehicle signals were first synchronized to a unified 50 Hz timebase. To maintain data integrity, continuous physical signals (e.g., speed) were aligned using linear interpolation, while discrete categorical data (e.g., ACC status) were mapped using nearest-neighbor interpolation.

    Following synchronization, overtaking events were manually annotated based on the four-phase overtaking model proposed by \cite{dozzaHowDriversOvertake2016}. We exclusively annotated the start ($t=0$) and end ($t_{end}$) times of the second phase (active passing phase), resulting in 223 total identified overtaking events. For each event, we extracted a Context Window ($[-5s, 0s]$), a 5-second pre-maneuver window aggregating kinematic and environmental state variables into the input feature vector $\mathbf{x}_i$, and an Action Window ($[0s, t_{end}]$), the duration of the active overtaking phase capturing the specific longitudinal acceleration profile $\mathbf{a}_i$ (cf. Fig. \ref{fig:framework}).
    
    To accommodate varying maneuver durations, action window data points were geometrically normalized to a fixed length for profile clustering and as the target output for sequence regression. In contrast, pre-maneuver context features were extracted as scalars from the raw context window for both intent classification and continuous residual regression.

    To strictly isolate actual driver expectations, we excluded all pure ACC events from Trip 1 and events where the driver did not override the ACC in Trip 3. Consequently, the final training dataset comprised $\text{N = 140}$ valid events of unconstrained manual driving (Trip 2) and shared control overrides (Trip 3).
    
    %%%%%%%%%%%%%%%%%%%%%%%%%%%%%%%%%%%%%%%%%%%%%%%%%%%%%%%%%%%%%%%%%%
	\section{Experiments and Results}
	\label{sec:resultsanddiscussion}

    The CoP-ACC framework is evaluated across three sequential dimensions: discrete classification accuracy, continuous profile fidelity, and distributional kinematic analysis.

    To evaluate the online prediction capabilities, the Random Forest Classifier was validated utilizing a 5-Fold Stratified Cross-Validation protocol with balanced class weighting. For generating the continuous expected acceleration, the 1D-CNN Decoder was evaluated on the computed residual datasets utilizing data augmentations (e.g., Gaussian Noise, Temporal Warping) to stabilize learning under limited human kinematic data. The resulting personalized profiles were validated through two complementary tests: Pairwise Profile Reconstruction Fidelity and Distributional Kinematic Analysis described in Section \ref{subsec:pairwise_profile_reconstruction} and \ref{subsec:distributional_validation}.

    \subsection{When and Why Drivers Override the ACC}

    Prior to generating personalized profiles, a statistical analysis of the pre-maneuver contexts examines why and when this individual rejects standard machine behavior. By analyzing experimental round trips where interventions were explicitly permitted, we identified sharp geometric boundaries that distinguish genuine ACC acceptance from bottleneck situations where inadequate acceleration prompts a manual override.
    \begin{figure}[!htbp]
        \centering
        \includegraphics[width=1\columnwidth]{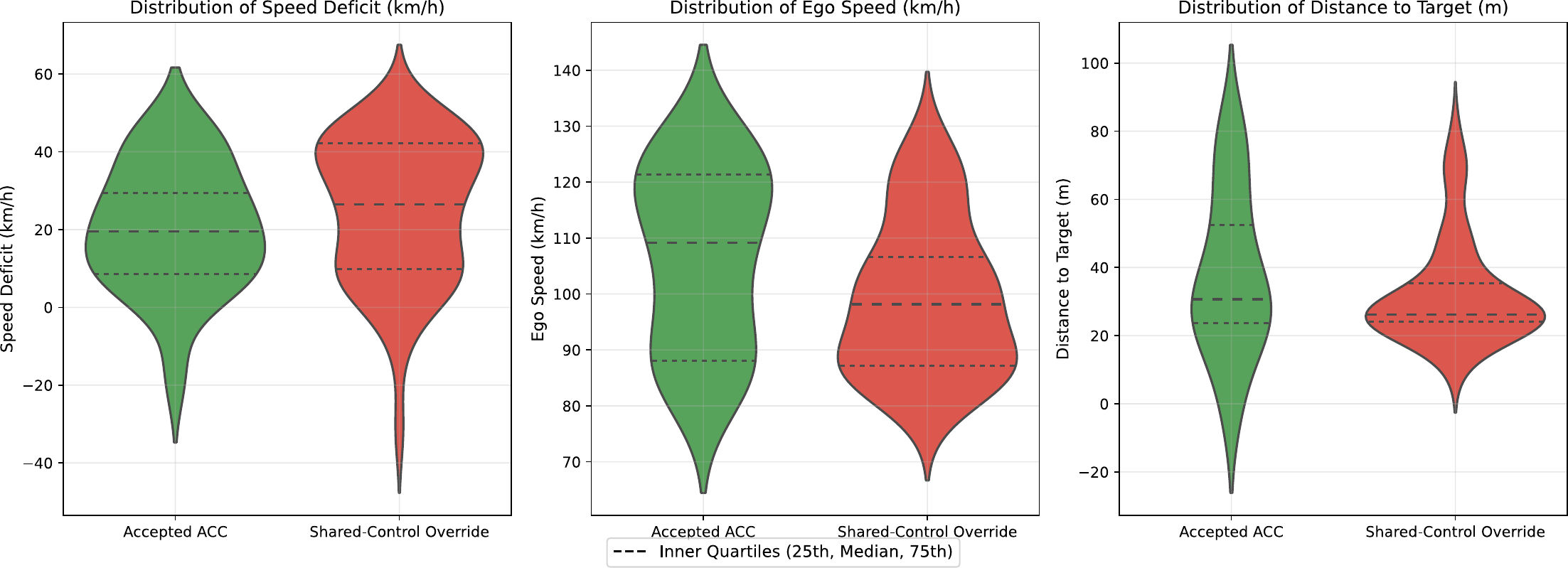}
        \caption{Distributional comparison of pre-maneuver kinematic features between ACC-accepted and shared control override events at overtake onset ($t=0$).}
        \label{fig:context_analysis}
    \end{figure}
    
    Fig. \ref{fig:context_analysis} compares the distributions of three pre-maneuver features between ACC-accepted and shared control override events. The left column shows that override events concentrate at notably higher speed deficits, meaning the driver intervenes when the ego vehicle is travelling well below the speed limit or its ACC target speed. The center column shows that overrides occur at lower ego speeds, reinforcing that the driver is stuck behind a slower vehicle. The right column reveals a tighter distance distribution shifted toward shorter gaps for override events, indicating the ACC's response becomes insufficient at close range. Together, these three observations suggest that the driver is more likely to override the standard ACC when constrained close behind a target with a large gap between current and desired speed, justifying the necessity of a context-driven prediction framework.
    
    \subsection{Acceleration Profile Clustering and Classification}

\begin{figure}[!htbp]
        \centering
        \includegraphics[width=1\columnwidth]{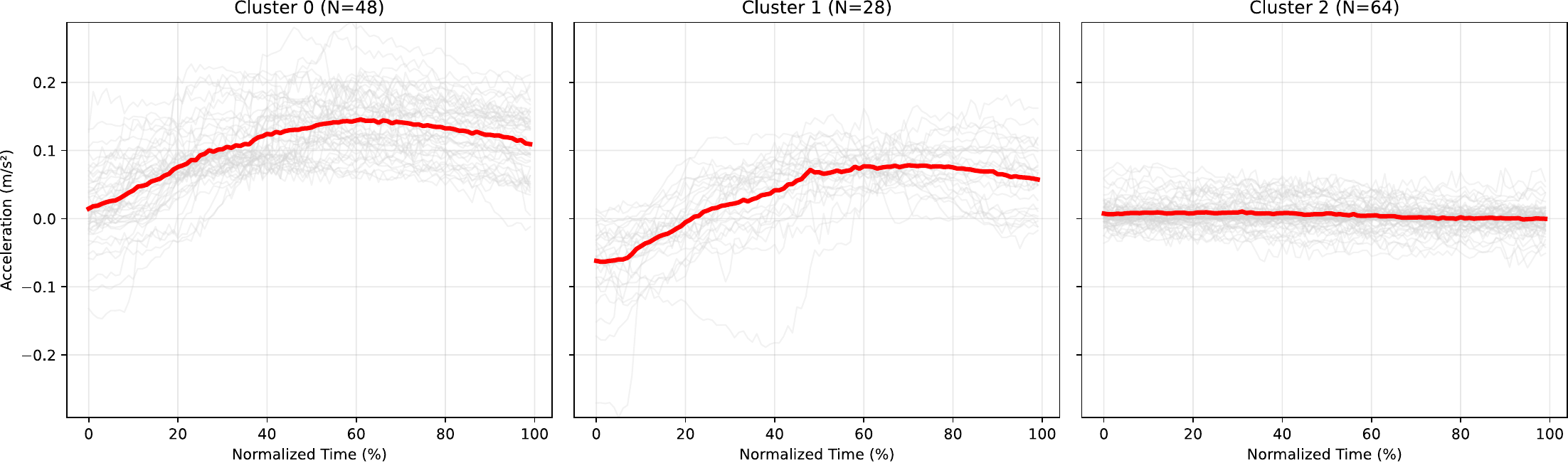}
        \caption{Longitudinal acceleration profiles for the three identified clusters ($K=3$). In red the mean, in light gray each individual profile.}
        \label{fig:cluster_profiles}
    \end{figure}

    The unsupervised hierarchical clustering algorithm identified three distinct empirical profiles as shown in Fig. \ref{fig:cluster_profiles}: Cluster 0 (aggressive, high-intensity acceleration peaking near $0.15 \text{ m/s}^2$), Cluster 1 (moderate, medium-intensity peaking near $0.08 \text{ m/s}^2$), and Cluster 2 (passive, near-zero flat profile at $0.01 \text{ m/s}^2$). Contextual correlations reveal that Clusters 0 and 1 correspond primarily to accelerative overtakes, where the ego vehicle initiates the pass from close behind the target and must actively accelerate to complete it. Conversely, Cluster 2 corresponds to flying overtakes, where the ego vehicle approaches with sufficient speed margin and passes at a comfortable distance, requiring minimal acceleration adjustment \cite{dozzaHowDriversOvertake2016, raschHowOncomingTraffic2020}.

    As detailed in Table \ref{tab:cluster_distribution}, the clustering separates human-machine conflict geometries. Remarkably, 64.4\% of all maneuvers assigned to Clusters 0 and 1 (aggressive and moderate acceleration) originated directly from shared control override events, indicating that these high-intensity profiles are primarily triggered when the individual must compensate for the generic ACC's insufficient acceleration.

    \begin{table}[!htbp]
    \centering
    \caption{Distribution of driving modes across the discrete clustering profiles}
    \label{tab:cluster_distribution}
    \resizebox{\columnwidth}{!}{%
    \begin{tabular}{|l|l|l|l|l|}
    \hline
    Event Driving Mode      & Cluster 0   & Cluster 1   & Cluster 2   & \textbf{Total} \\ \hline
    Naturalistic Manual     & 19          & 8           & 61          & \textbf{88}    \\ \hline
    Shared-Control Override & 29          & 20          & 3           & \textbf{52}    \\ \hline
    \textbf{Total}          & \textbf{48} & \textbf{28} & \textbf{64} & \textbf{140}   \\ \hline
    \end{tabular}%
    }
    \end{table}
    
\begin{figure}[!htbp]
        \centering
        \includegraphics[width=0.78\columnwidth]{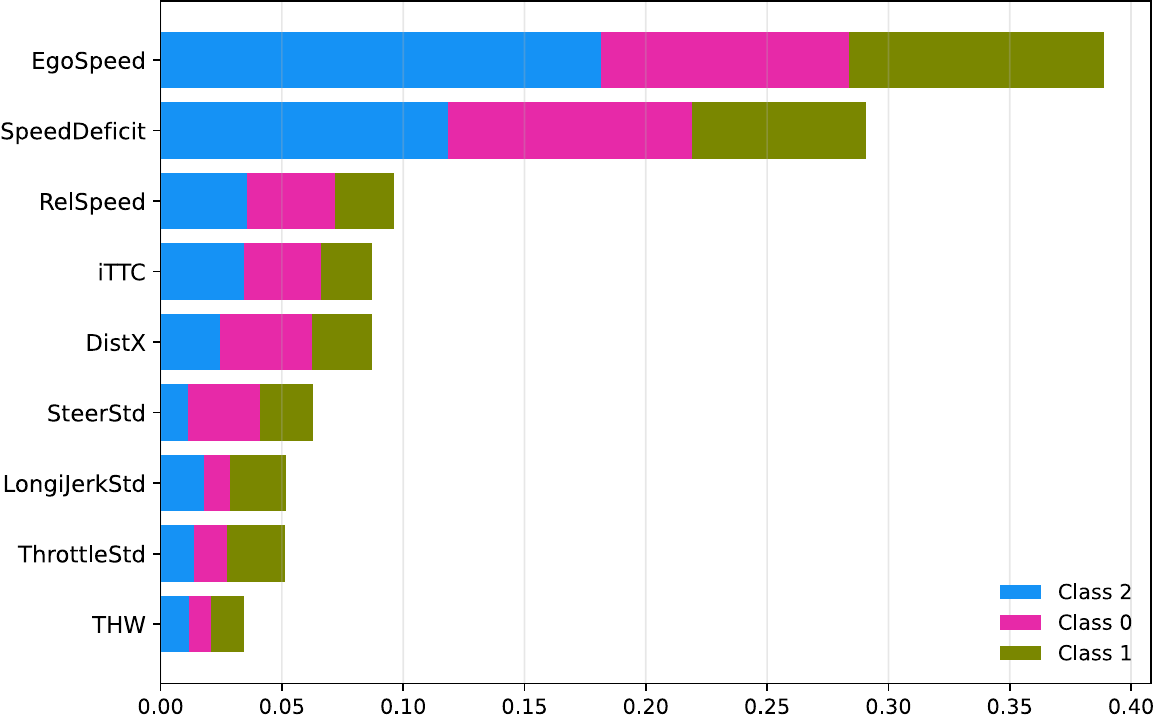}
        \caption{SHAP global feature importance for Random Forest cluster classification.}
        \label{fig:shap_importance}
    \end{figure}

    Feature importance analysis evaluated via SHAP in Fig. \ref{fig:shap_importance} revealed that kinematic motivation variables, specifically speed deficit and ego speed, dominated the cluster prediction. In contrast, behavioral style variables had negligible predictive influence, indicating that the acceleration profile of this individual driver primarily depends on specific kinematic tolerance thresholds.
    
    Exploiting these kinematic boundaries, the Random Forest classifier achieved an accuracy of 89.29\% on the test set, with macro precision of 85.19\%, macro recall of 86.67\%, and macro F1-score of 85.65\%.

    \subsection{Pairwise Profile Reconstruction Fidelity}
    \label{subsec:pairwise_profile_reconstruction}

    This test evaluates whether the predicted profile $\hat{\mathbf{a}}_i = \mathbf{a}^*_{\hat{y}} + \hat{\mathbf{r}}$ can reconstruct the driver's actual acceleration $\mathbf{a}_i$ at each of the $L=100$ normalized sample points.
    We isolated the shared control override events, fed their contexts into the pipeline (cf Fig. \ref{fig:framework}), and directly overlaid the predicted output onto the raw human profile. Fidelity was quantified through three complementary metrics:
    \begin{equation}
    \text{Mean Absolute Error (MAE)} = \frac{1}{L}\sum_{l=1}^{L} |a_{i,l} - \hat{a}_{i,l}|
    \label{eq:mae}
    \end{equation}
    \noindent MAE measures the average point-wise longitudinal acceleration error ($\text{m/s}^2$). Dynamic Time Warping (DTW) distance \cite{sakoeDynamicProgrammingAlgorithm1978} quantifies shape similarity under temporal misalignment by accumulating point-wise acceleration differences along the normalized profiles ($\text{m/s}^2$,). Pearson Correlation Coefficient $r(\mathbf{a}_i, \hat{\mathbf{a}}_i)$ \cite{benestyPearsonCorrelationCoefficient2009} is unitless and evaluates temporal trend alignment between the two profiles.

    Integrating the output of the classifier with the conditional residual regressor yields a continuous profile parameterization. The Context-Conditioned 1D-CNN Decoder improved over the centroid-only baseline, reducing Mean Squared Error (MSE) from 0.001308 $\text{m/s}^2$ to 0.001088 $\text{m/s}^2$ and MAE from 0.02522 $\text{m/s}^2$ to 0.02398 $\text{m/s}^2$.
    
    \begin{figure}[!htbp]
        \centering
        \includegraphics[width=1\columnwidth]{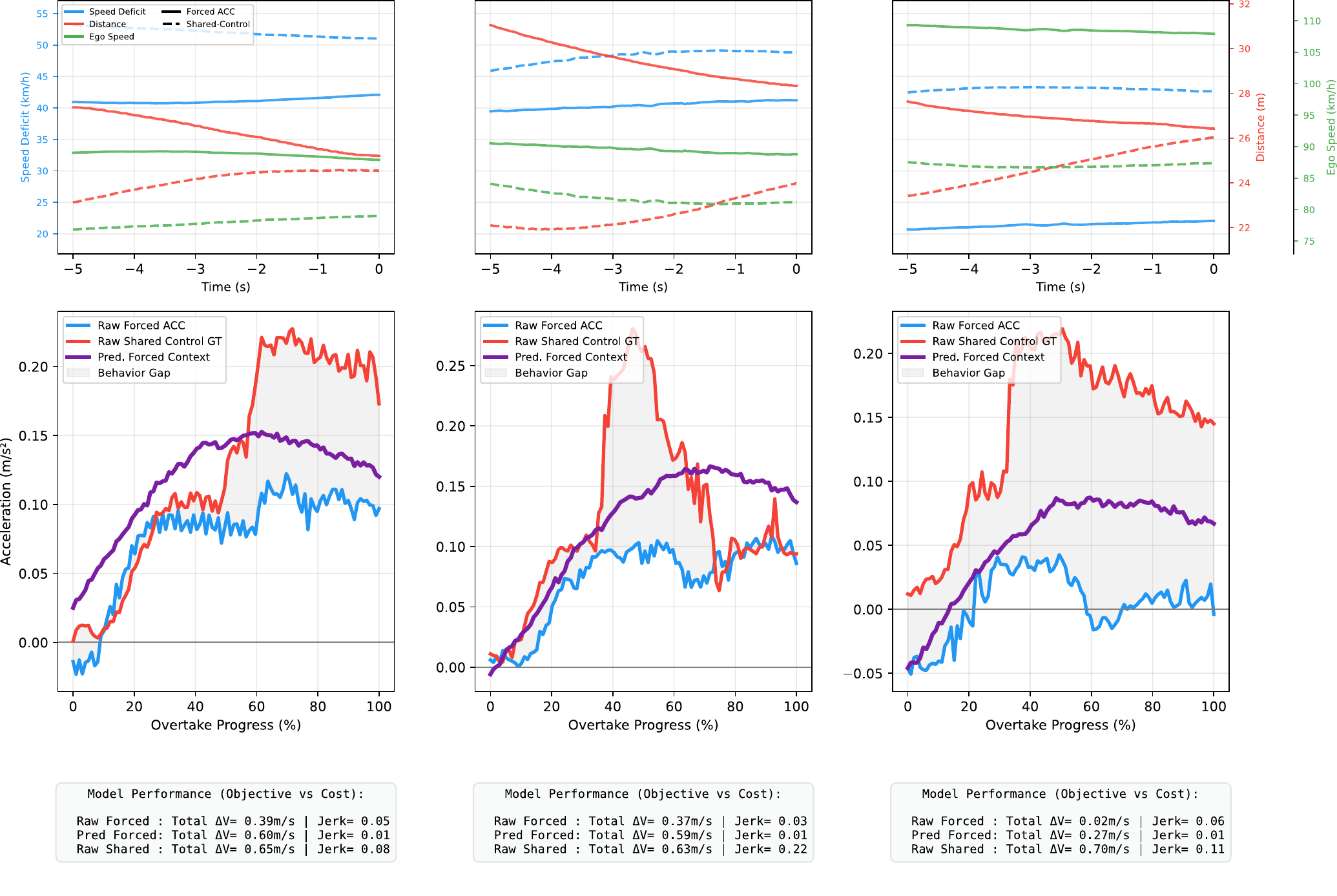}
        \caption{Context-matched comparison of generic ACC, shared control override, and predicted acceleration profiles with corresponding kinematic metrics.}
        \label{fig:counterfactual_matched}
    \end{figure}

    We extract three diverse pairs of events with similar driving contexts from forced ACC events and shared control override events. As shown in Fig. \ref{fig:counterfactual_matched}, the first row shows the contextual similarity between each pair. The second row reveals the central result: given an ACC baseline context, the predicted profile clearly departs from the conservative, delayed ACC acceleration and converges toward the shared control override profile observed in the matching context. This indicates that the CoP-ACC framework can produce, from a context where the driver had no control, an acceleration profile close to what the driver would have naturally demanded. The third row further shows that the predicted profiles achieve nearly identical AUC to the human override while generating significantly lower jerk, indicating a smoother execution of the same driving intent.

    For the shared control override test subset, the integrated pipeline reconstructed the driver's acceleration trend with an MAE of $0.0306 \text{ m/s}^2$, an average DTW Distance of $2.00$, and an average Pearson Correlation Coefficient of $r=0.804$. These metrics indicate that the predicted acceleration remains close to the human pedal input while preserving the main temporal shape of the driver's intended response.

    \subsection{Distributional Validation of Personalization}
    \label{subsec:distributional_validation}

    While pairwise fidelity evaluates isolated events, a secondary validation analyzes the resulting kinematic distribution on withheld forced-ACC contexts to confirm system generalization. To achieve this, we structured an independent evaluation comparing three distinct groups of acceleration profiles: Actual Overrides, i.e., the raw profiles physically executed by the human driver during shared control override events; Machine Baseline, i.e., the raw profiles executed by the generic ACC during ``potential override" events extracted from Trip 1 (where the driver was forbidden from overriding), restricted to contexts that mathematically match the thresholds of actual override events; and Predicted, i.e., the synthetic profiles generated by our framework when conditioned on the ``potential override" contexts of Machine Baseline.
   
    To quantify system performance, we extracted core integral and differential kinematic features from each trajectory: Area Under the Curve (AUC) of Total Velocity Gained, Peak Longitudinal Acceleration, and Maximum Jerk.

\begin{figure}[!htbp]
        \centering
        \includegraphics[width=1\columnwidth]{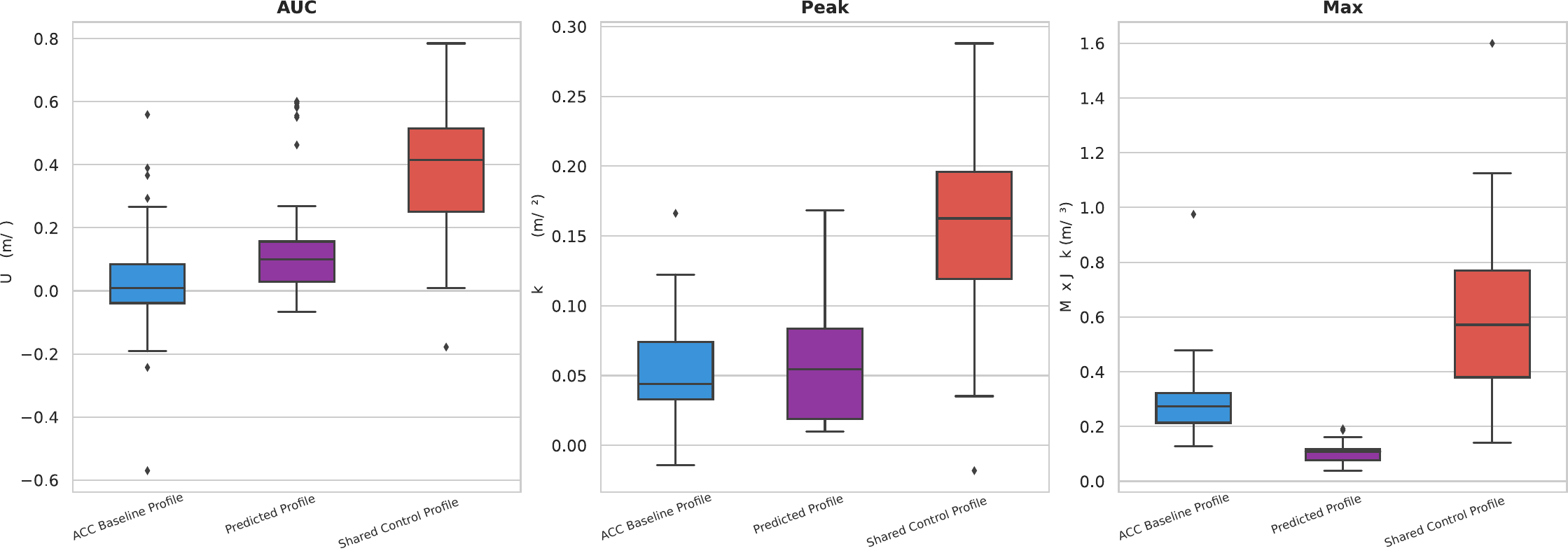}
        \caption{Distributional comparison of kinematic features (AUC, Peak Acceleration, Maximum Jerk) across machine baseline, actual overrides, and predicted profiles.}
        \label{fig:distributional_boxplots}
    \end{figure}

    Applying the continuous prediction pipeline to all ``potential override" events illustrates the main advantage of our system: it shifts the car's acceleration toward the driver's observed preference while keeping the ride smooth. As demonstrated in the distributional boxplots of Fig. \ref{fig:distributional_boxplots}, evaluating the generated profiles column-by-column reveals the mechanism of this personalization.

    First, examining the AUC of Total Velocity Gained reveals that the predicted profiles capture the driver's macro-objective. The predictive framework elevates the AUC distribution out of the congested, slow baseline of the generic ACC to closely match the broader, higher-velocity distribution of the human expectation, indicating that the identified model initiates the early acceleration demanded by the driver.
    
    Second, examining the Peak Acceleration and Maximum Jerk reveals a comfort-oriented effect. The predicted profiles exhibit a lower overall peak acceleration compared to the human override. This occurs because the predictive framework anticipates the maneuver and initiates acceleration earlier than the generic ACC (as shown in AUC), reducing the need for the late, hard acceleration that the driver executes during intervention. Importantly, the predicted profiles maintain a notably higher average peak acceleration than the sluggish machine baseline, indicating that the system preserves the required agility. Finally, the predicted profiles achieve a maximum jerk distribution that is lower than both the human override and the baseline ACC profiles, suggesting a smoother driving experience.
	
    %%%%%%%%%%%%%%%%%%%%%%%%%%%%%%%%%%%%%%%%%%%%%%%%%%%%%%%%%%%%%%%%%%
	% \section{Proposed Simulation Validation}
	% \label{sec:simulation}
	
 %    To validate closed-loop safety and real-time operational performance, we propose a co-simulation evaluation framework utilizing IPG CarMaker. Our trained contextual classifier and continuous residual regressor will be integrated to dynamically recreate varied naturalistic overtaking scenarios.
	
 %    Mirroring the offline experimental evaluation structure, three distinct trajectory executions will be compared within each simulated context: the baseline generic ACC, the human driver's ground-truth shared control override profile, and the synthesized acceleration profile generated by proposed framework.

 %    Validation relies on dynamic trajectory convergence across two distinct scenarios, evaluated both qualitatively and quantitatively. First, in "potential override" contexts (high-urgency scenarios where generic ACC is overly conservative), the framework is successful if the predicted acceleration dynamically converges toward the shared control override profile. Second, in "accepted ACC" scenarios (low-urgency contexts where the driver historically accepted the baseline behavior), the predicted profile must remain strictly aligned with the passive generic ACC baseline. This dual-scenario evaluation ensures the framework actively personalizes aggressive maneuvers when required, without inducing uncomfortable or unsafe accelerations in benign contexts.
    
    %%%%%%%%%%%%%%%%%%%%%%%%%%%%%%%%%%%%%%%%%%%%%%%%%%%%%%%%%%%%%%%%%%
	\section{Conclusion and Future Work}
	\label{sec:conclusion}
	
    This work demonstrates the feasibility of reframing manual pedal interventions as shared-control override signals, enabling a human-in-the-loop learning paradigm for personalized ACC adaptation. The proposed CoP-ACC hybrid framework extracts distinct acceleration profiles reflecting driver urgency and maps pre-maneuver contexts to smooth, personalized acceleration profiles through a Random Forest context classifier coupled with a 1D-CNN Decoder. Counterfactual evaluation indicates the intended behavioral shift: when conditioned on baseline ACC contexts associated with potential intervention, the predicted profiles tend toward the acceleration patterns observed during real override events.

    All data were collected from production vehicles operating on public roads, grounding the results in real-world driving behavior. Despite being trained on a constrained single-driver dataset ($\text{N = 140}$), CoP-ACC achieves low reconstruction error, strong temporal correlation with actual driver inputs, and distributional alignment with human override behavior. These results indicate that meaningful behavioral adaptation can emerge even at limited scale. Nevertheless, generalization across heterogeneous populations remains to be validated. Current feature set excludes road geometry (e.g., curvature), and the dataset required augmentation to stabilize learning, both aspects motivating broader real-world data collection.

    A closed-loop simulator study has been initiated to evaluate behavioral adaptation under interactive conditions. Future work will focus on translating predicted acceleration profiles into adaptive ACC parameters (e.g., time gap, response dynamics) using learning-based optimization approaches, while preserving the existing ACC architecture and safety barriers. Scalable deployment will also require automatic event detection, and validating CoP-ACC with diverse drivers in instrumented vehicles. The ultimate objective is anticipatory ACC behavior that reduces manual overrides while preserving driver comfort and intent alignment.

    %%%%%%%%%%%%%%%%%%%%%%%%%%%%%%%%%%%%%%%%%%%%%%%%%%%%%%%%%%%%%%%%%%
    \section*{Acknowledgments}
    \label{ack}

    This work was supported in part by Renault Group and ANRT through a CIFRE (Convention Industrielle de Formation par la Recherche) and in part by the French Government, through the CPER RITMEA, Hauts-de-France Region. This work has been also partially supported by ROBOTEX 2.0, funded by the French program France 2030.
    
    %%%%%%%%%%%%%%%%%%%%%%%%%%%%%%%%%%%%%%%%%%%%%%%%%%%%%%%%%%%%%%%%%%
	%\addtolength{\textheight}{-12cm}
	%\vspace{10mm}
	\bibliographystyle{IEEEtran}
	% Your .bib file here
	\bibliography{root} 
	
\end{document}